\documentclass[runningheads]{llncs}
\usepackage[T1]{fontenc}
\usepackage{graphicx}
\usepackage{amsfonts}
\usepackage{amsmath}
\usepackage{multirow}
\usepackage{booktabs}
\usepackage{url}
\begin{document}
\title{Towards Surgical World-Action Modeling: A Preliminary Joint Visual-Trajectory Forecasting for Surgical Motion Planning}%
%
\author{Weiliang Huang\inst{1} \and
Huanrong Liu\inst{1,3} \and
Bob Zhang\inst{1}$^{,\dagger}$\and
Qi Dou\inst{4} \and
Zhen Chen\inst{5} \and
Yun Gu\inst{6,7} \and
Guy Rosman\inst{8} \and
Qingbiao Li\inst{1,2,3}$^{,\dagger}$
}

\authorrunning{W. Huang et al.}
%
\institute{Faculty of Information Science and Computing, University of Macau, Macau, China \and
Faculty of Engineering, University of Macau, Macau, China \and
University of Macau Advanced Research Institute in Hengqin, Zhuhai, China \and
Department of Computer Science and Engineering, The Chinese University of Hong Kong, Hong Kong, China \and
Department of Data Science and Artificial Intelligence, The Hong Kong Polytechnic University, Hong Kong, China \and
Shanghai Key Laboratory of Flexible Medical Robotics, Tongren Hospital, Institute of Medical Robotics, Shanghai Jiao Tong University, Shanghai, China \and
School of Automation and Intelligent Sensing, Shanghai Jiao Tong University, Shanghai, China \and
School of Medicine, Duke University, Durham, North Carolina, USA 
\email{qingbiaoli@um.edu.mo}\\
\quad$^{\dagger}$\,Corresponding author}
\maketitle  
\begin{abstract}
Reliable surgical planning requires models to anticipate not only how instruments will move, but also how the operative visual state will evolve together with such motion. Existing approaches typically treat future scene generation and instrument trajectory prediction as two separate tasks. Scene-only models cannot directly evaluate the accuracy of future instrument motion at the trajectory level, while trajectory-only models fail to capture the visual consequences of instrument movement, leaving the consistency between predicted trajectories and future scene evolution unaddressed. Jointly forecasting both provides a more complete account of surgical action–scene dynamics by enabling explicit trajectory-level evaluation while simultaneously modeling the corresponding visual evolution. To bridge this gap, we present a preliminary joint visual-trajectory world-action model that simultaneously forecasts future visual states and instrument trajectories from historical surgical observations. Specifically, we encode historical video frames and tool trajectories into latent representations, which are processed by a temporal-spatial encoder and subsequently decoded through separate visual-state and trajectory prediction heads. Based on this preliminary architecture, a chunked autoregressive rollout is repeatedly applied to predict fifteen future steps. The chunked strategy consistently outperforms direct one-shot prediction across all evaluated horizons, improving first-segment PSNR from 18.86 to 23.11\,dB and reducing ADE from 45.77 to 22.22 pixels. These results demonstrate the initial feasibility of joint visual-motion forecasting. However, we observe progressive visual degradation and accumulated trajectory errors over longer prediction horizons, which remain important challenges for future surgical world-action modeling.

\end{abstract}
\section{Introduction}
Reliable surgical planning \cite{attanasio2021autonomy,saeidi2022autonomous} requires more than estimating where an instrument will move next. Surgical actions continuously reshape the operative field through instrument–tissue interactions, tissue deformation, occlusion, bleeding, and changes in camera visibility. Consequently, a predicted trajectory may be geometrically close to an expert demonstration while still leading to an implausible or unsafe future surgical state. A planning model should therefore anticipate not only the future location of the instrument, but also how the local scene will change. However, current surgical trajectory prediction and future visual state prediction have largely developed as separate research directions. 

Trajectory methods~\cite{shi2022recognition,hansen2026imitatecholec,liu2026sutureagent} typically forecast two-dimensional coordinates, robotic states, or motion commands without explicitly representing their visual consequences. Conversely, surgical video generation methods~\cite{biagini2025hierasurg,chen2025surgsora,cho2024surgen} model future appearance and scene dynamics but often provide limited evidence that the generated motion is geometrically accurate. This separation limits the ability of existing methods to jointly reason about instrument motion and the corresponding evolution of the operative environment, which is a core requirement for surgical world modeling and closed-loop planning.

Recent surgical world models \cite{koju2025surgicalvision,chen2025surgsora,rapuri2026saw,he2025surgworld} have demonstrated promising capabilities in future video generation. Their evaluation, however, remains predominantly focused on generation-oriented criteria such as visual realism, temporal consistency, Fréchet Video Distance \cite{unterthiner2018towards}, PSNR, and SSIM. Although these metrics are useful for assessing appearance similarity, visually plausible predictions do not necessarily imply accurate instrument motion or stable long-horizon dynamics. A generated sequence may preserve the overall appearance of the surgical scene while exhibiting endpoint deviation, incorrect local motion, or progressively increasing drift during autoregressive rollout. These limitations motivate a world-action formulation that extends conventional future-state prediction by explicitly coupling world evolution with action-relevant motion. In our surgical setting, the future visual representation describes how the operative scene evolves, while the instrument trajectory provides an explicit representation of tool motion. Jointly predicting these complementary outputs enables visual-state evolution and instrument motion to be modeled and evaluated within a unified predictive framework, providing a more complete characterization of surgical dynamics.

In this work, we preliminarily investigate whether a predictive model can jointly forecast future visual representations and instrument trajectories from the same historical surgical 
information. Our model first encodes the observed surgical frames and historical tool trajectories into latent representations. These representations are then processed by a temporal-spatial encoder and fed into separate heads for future visual-state prediction and two-dimensional instrument trajectory prediction. To support long-horizon prediction, the model generates three future steps at each stage and recursively feeds the predicted visual and motion representations back into the model, producing a fifteen-step rollout through five successive prediction chunks. Scheduled sampling \cite{bengio2015scheduled} is further employed during training to gradually expose the model to its own predictions and partially reduce the discrepancy between training and autoregressive inference. We evaluate the proposed baseline on SurgWMBench~\cite{liu2026surgwmbench}, a benchmark for surgical motion planning using endoscopic video sequences and annotated two-dimensional instrument trajectories. Our contributions are threefold:

1. We introduce a  joint predictive model that simultaneously forecasts future visual latent representations and two-dimensional surgical instrument trajectories.

2. We develop a chunked autoregressive strategy that decomposes the fifteen-step prediction horizon into repeated three-step rollouts, while scheduled sampling partially reduces the train--inference discrepancy and improves long-horizon visual-motion stability.

3. We provide a preliminary evaluation on SurgWMBench, demonstrating the feasibility of joint visual-trajectory forecasting while highlighting remaining challenges in long-horizon visual fidelity and trajectory stability.

\section{Method}
\subsection{Problem Formulation}
Given a sequence of $K$ observed surgical frames,
\begin{equation}
    \mathcal{I}_{1:K}
    =
    \left\{
    I_1,\ldots,I_K
    \right\},
\end{equation}
and the corresponding historical two-dimensional instrument trajectory,
\begin{equation}
    \mathcal{P}_{1:K}
    =
    \left\{
    \mathbf{p}_1,\ldots,\mathbf{p}_K
    \right\},
    \qquad
    \mathbf{p}_t \in \mathbb{R}^{2},
\end{equation}
our objective is to jointly predict future visual states and instrument
trajectories over a horizon of $H$ steps.

Instead of directly predicting future frames in the pixel space, we represent
each future visual state using a latent feature representation. The joint
prediction task is formulated as
\begin{equation}
    \left(
    \hat{\mathcal{Z}}_{K+1:K+H},
    \hat{\mathcal{P}}_{K+1:K+H}
    \right)
    =
    F_{\theta}
    \left(
    \mathcal{I}_{1:K},
    \mathcal{P}_{1:K}
    \right),
    \label{eq:joint_prediction}
\end{equation}
where $\hat{\mathcal{Z}}_{K+1:K+H}$ denotes the predicted future
visual representations and $\hat{\mathcal{P}}_{K+1:K+H}$ denotes the
predicted instrument trajectory. A frozen visual decoder $D$ maps each
predicted latent representation back to an RGB frame for visualization and
image-level evaluation:
\begin{equation}
    \hat{I}_t = D\left(\hat{Z}_t\right).
    \label{eq:visual_decoder}
\end{equation}

In this preliminary formulation, the future instrument trajectory is treated
as an action-oriented planning representation rather than an explicit
low-level robotic control command.

\begin{figure}[t]
    \centering
    \includegraphics[width=\linewidth]
    {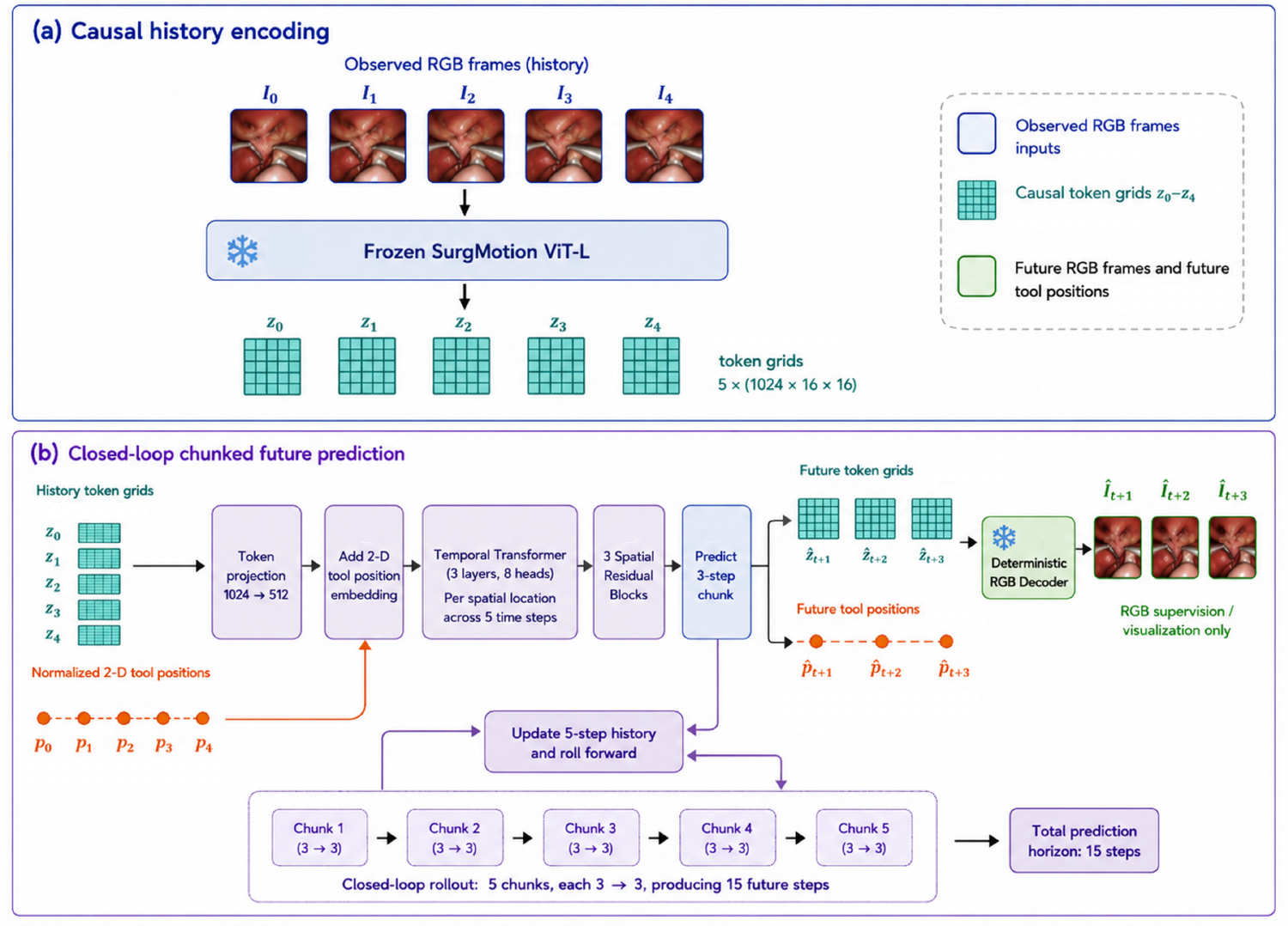}
    \caption{Overview of the proposed method. (a) Causal history encoding: five observed RGB frames are processed by the frozen SurgMotion ViT-L encoder to obtain causal token grids. (b) Closed-loop chunked future prediction: the predictor jointly forecasts future token grids and normalized 2-D tool positions in 3-step chunks; each predicted chunk updates the 5-step history for the next rollout, producing a total 15-step forecast.}
    \label{fig:framwork}
\end{figure}

\subsection{Joint Visual-Motion Representation}
The historical surgical frames are first processed by a frozen SurgMotion \cite{wu2026unisurg} encoder as shown in Figure \ref{fig:framwork}. We extract intermediate-layer visual tokens that retain local appearance and spatial details, together with final-layer visual tokens that capture higher-level semantic and motion context. The two visual feature streams are independently projected into a shared hidden dimension. For each historical step, the corresponding two-dimensional trajectory coordinate $\mathbf{p}_t$ is embedded using a multilayer perceptron. The resulting trajectory embedding is spatially broadcast and fused with the visual representations:
\begin{equation}
    X_t
    =
    \phi_{\mathrm{mid}}
    \left(
    Z_t^{\mathrm{mid}}
    \right)
    +
    \phi_{\mathrm{final}}
    \left(
    Z_t^{\mathrm{final}}
    \right)
    +
    \phi_{\mathrm{traj}}
    \left(
    \mathbf{p}_t
    \right),
    \label{eq:feature_fusion}
\end{equation}
where $\phi_{\mathrm{mid}}$, $\phi_{\mathrm{final}}$, and
$\phi_{\mathrm{traj}}$ denote the corresponding projection modules. This
fusion constructs a shared representation containing both operative-scene
information and historical instrument motion.

A temporal-spatial encoder is subsequently employed to model the evolution
of the fused visual-motion representations. Temporal attention captures the
dependencies among historical states, while spatial processing preserves the
local structure of the operative field. The encoded feature corresponding to
the latest historical state is used as the shared context representation
$C$. Learned future-step embeddings are then combined with $C$ to generate
step-specific representations for future prediction.

Two separate prediction heads operate on the shared context. The visual-state
prediction head estimates a residual change with respect to the latest
observed visual representation:

\begin{equation}
    \hat{Z}_{K+j}
    =
    Z_K^{\mathrm{}}
    +
    \Delta Z_{K+j},
    \qquad
    j=1,\ldots,c,
    \label{eq:visual_residual}
\end{equation}
where $c=3$ denotes the prediction chunk length. Predicting residual visual
changes encourages the model to preserve the observed scene structure while
modeling future state transitions.

For trajectory prediction, the shared context is spatially pooled and
combined with the latest instrument position and motion information. We first
construct a constant-velocity estimate:
\begin{equation}
    \tilde{\mathbf{p}}_{K+j}
    =
    \mathbf{p}_K
    +
    j
    \left(
    \mathbf{p}_K-\mathbf{p}_{K-1}
    \right).
    \label{eq:constant_velocity}
\end{equation}
The trajectory prediction head then estimates a residual correction
$\mathbf{r}_{K+j}$:
\begin{equation}
    \hat{\mathbf{p}}_{K+j}
    =
    \tilde{\mathbf{p}}_{K+j}
    +
    \mathbf{r}_{K+j}.
    \label{eq:trajectory_residual}
\end{equation}
This formulation introduces a simple motion prior while allowing the learned
visual-motion representation to correct the trajectory direction, magnitude,
and endpoint.

\subsection{Chunked Autoregressive Rollout}
Directly predicting all fifteen future steps in a single forward may produce temporally averaged visual states or inaccurate long-range trajectories. We therefore decompose the
full prediction horizon into several short prediction chunks.

Given a five-step visual-motion history, the model predicts the next
$c=3$ visual representations and trajectory points:
\begin{equation}
    \left(
    \hat{\mathcal{Z}}_{t+1:t+3},
    \hat{\mathcal{P}}_{t+1:t+3}
    \right)
    =
    F_{\theta}
    \left(
    \mathcal{Z}_{t-4:t},
    \mathcal{P}_{t-4:t}
    \right).
    \label{eq:chunk_prediction}
\end{equation}

After each prediction stage, the oldest three historical states are removed,
and the three predicted visual representations and trajectory points are
appended to construct a new five-step history window. The updated history is
then used as the input to the next prediction stage. 

During training, scheduled sampling is employed to reduce the discrepancy
between teacher-forced training and fully autoregressive inference. At each
rollout stage, the history input is constructed using either a ground-truth
state or a model prediction:
\begin{equation}
    \bar{x}_t
    =
    m_t \hat{x}_t
    +
    \left(1-m_t\right)x_t,
    \qquad
    m_t
    \sim
    \operatorname{Bernoulli}\left(\rho_e\right),
    \label{eq:scheduled_sampling}
\end{equation}
where $x_t$ and $\hat{x}_t$ denote the ground truth and predicted states,
respectively. The variable $\rho_e$ denotes the probability of selecting a
predicted state at training epoch $e$. This probability gradually increases
during training, exposing the model to its own prediction errors while
maintaining stable supervision during the early training stages. During
inference, all prediction chunks are generated entirely from the model's
previous predictions.

\section{Experiments}

\subsection{Dataset and Evaluation Protocol}
We conduct experiments on SurgWMBench~\cite{liu2026surgwmbench}, a vision-based surgical motion planning benchmark constructed from SAR-RARP50~\cite{psychogyios2023sar}, which contains real clinical videos of Robot-Assisted Radical Prostatectomy (RARP) performed using the da Vinci surgical system. SurgWMBench focuses on suturing motions and extracts effective needle insertion and extraction actions as individual motion segments. The benchmark contains 1,637 valid segments, corresponding to 32,740 annotated surgical frames. For each frame, one task-relevant two-dimensional instrument anchor point is manually annotated, yielding a 20-point trajectory for each segment. Following the short-horizon prediction protocol of SurgWMBench~\cite{liu2026surgwmbench}, we use the first five frames and their corresponding trajectory points as historical observations and predict the remaining fifteen future visual states and trajectory points. Trajectory errors are evaluated in the original image pixel space.

We compare two prediction settings under the same strict history-only protocol. 
In the \emph{direct one-shot} setting, the model predicts all fifteen future 
visual representations and trajectory points in a single forward pass. In the 
\emph{chunked rollout} setting, the model predicts three future steps at each 
stage and recursively uses the predicted visual-motion states to construct the 
input for the next stage. This process is repeated five times to obtain the 
complete fifteen-step prediction.

For visual-state evaluation, the predicted latent representations are mapped 
to RGB frames using the frozen decoder. We report Peak Signal-to-Noise Ratio 
(PSNR) \cite{huynh2008scope}, Structural Similarity Index Measure (SSIM) \cite{wang2004image}, and Learned Perceptual Image Patch Similarity (LPIPS) \cite{zhang2018unreasonable}. Higher PSNR and SSIM indicate better visual 
reconstruction, whereas lower LPIPS indicates stronger perceptual similarity. 
For trajectory evaluation, we report Average Displacement Error (ADE) and 
Final Displacement Error (FDE) \cite{gupta2018social} in the original image pixel space. ADE measures 
the average point-wise trajectory error, while
FDE measures the displacement error at the final point of that segment.

All metrics are reported separately over five non-overlapping future segments: $t+1{:}3$, $t+4{:}6$, $t+7{:}9$, $t+10{:}12$, and $t+13{:}15$, where $t$ denotes the current frame and all prediction offsets are specified in frames.
This segment-wise evaluation allows us to examine how visual quality and 
trajectory accuracy change as the prediction horizon increases.

\subsection{Implementation Details}
We train with AdamW ($7\times10^{-5}$ learning rate and $10^{-4}$ weight decay), a five-epoch warm-up, cosine decay to $5\%$ of the initial rate, batch size one, and early stopping with patience 14 (maximum 70 epochs). The loss combines token MSE/cosine, RGB, Sobel-edge, temporal-difference, trajectory-position, velocity, and endpoint losses with weights $1$, $0.05$, $0.25$, $0.10$, $0.10$, $5$, $2$, and $2$, respectively; step weights increase from 1.0 to 1.8 over the future horizon. Scheduled sampling starts at epoch 8 and reaches a 0.75 predicted-state probability at epoch 45; validation and test are fully closed loop. All training experiments were performed on a single NVIDIA RTX 5090 GPU.

\subsection{Quantitative Results}
Table~\ref{tab:quantitative_results} compares the direct one-shot prediction 
and chunked autoregressive rollout settings. The chunked rollout consistently 
outperforms direct prediction across all five future segments and all evaluated 
metrics.

\begin{table}[t]
    \centering
    \caption{Segment-wise comparison between chunked autoregressive rollout 
    and direct one-shot prediction. Higher values are better for PSNR and 
    SSIM, while lower values are better for LPIPS, ADE, and FDE. Trajectory 
    errors are measured in pixels.}
    \label{tab:quantitative_results}
    \small
    \setlength{\tabcolsep}{4.5pt}
    \begin{tabular}{ccccccc}
        \toprule
        & & \multicolumn{3}{c}{Visual Metrics}
        & \multicolumn{2}{c}{Trajectory Metrics} \\
        \cmidrule(lr){3-5}
        \cmidrule(lr){6-7}
        Setting
        & Segment
        & PSNR $\uparrow$
        & SSIM $\uparrow$
        & LPIPS $\downarrow$
        & ADE $\downarrow$
        & FDE $\downarrow$ \\
        \midrule

        \multirow{5}{*}{\shortstack[c]{Chunked\\rollout}}
        & $t+1{:}3$
        & 23.105 & 0.7971 & 0.1783
        & 22.22 & 31.64 \\
        & $t+4{:}6$
        & 20.395 & 0.7187 & 0.2380
        & 52.78 & 62.64 \\
        & $t+7{:}9$
        & 19.254 & 0.6799 & 0.2700
        & 84.13 & 94.27 \\
        & $t+10{:}12$
        & 18.495 & 0.6527 & 0.2906
        & 119.38 & 133.06 \\
        & $t+13{:}15$
        & 17.982 & 0.6343 & 0.3047
        & 158.52 & 172.12 \\
        \midrule

        \multirow{5}{*}{\shortstack[c]{Direct one-shot\\prediction}}
        & $t+1{:}3$
        & 18.859 & 0.4713 & 0.6045
        & 45.77 & 49.47 \\
        & $t+4{:}6$
        & 17.963 & 0.4482 & 0.6138
        & 72.58 & 85.47 \\
        & $t+7{:}9$
        & 17.489 & 0.4385 & 0.6182
        & 119.01 & 132.94 \\
        & $t+10{:}12$
        & 17.132 & 0.4314 & 0.6221
        & 156.11 & 169.57 \\
        & $t+13{:}15$
        & 16.872 & 0.4264 & 0.6244
        & 195.27 & 208.04 \\
        \bottomrule
    \end{tabular}
\end{table}

\begin{figure}[t]
    \centering
    \includegraphics[width=0.9\linewidth]
    {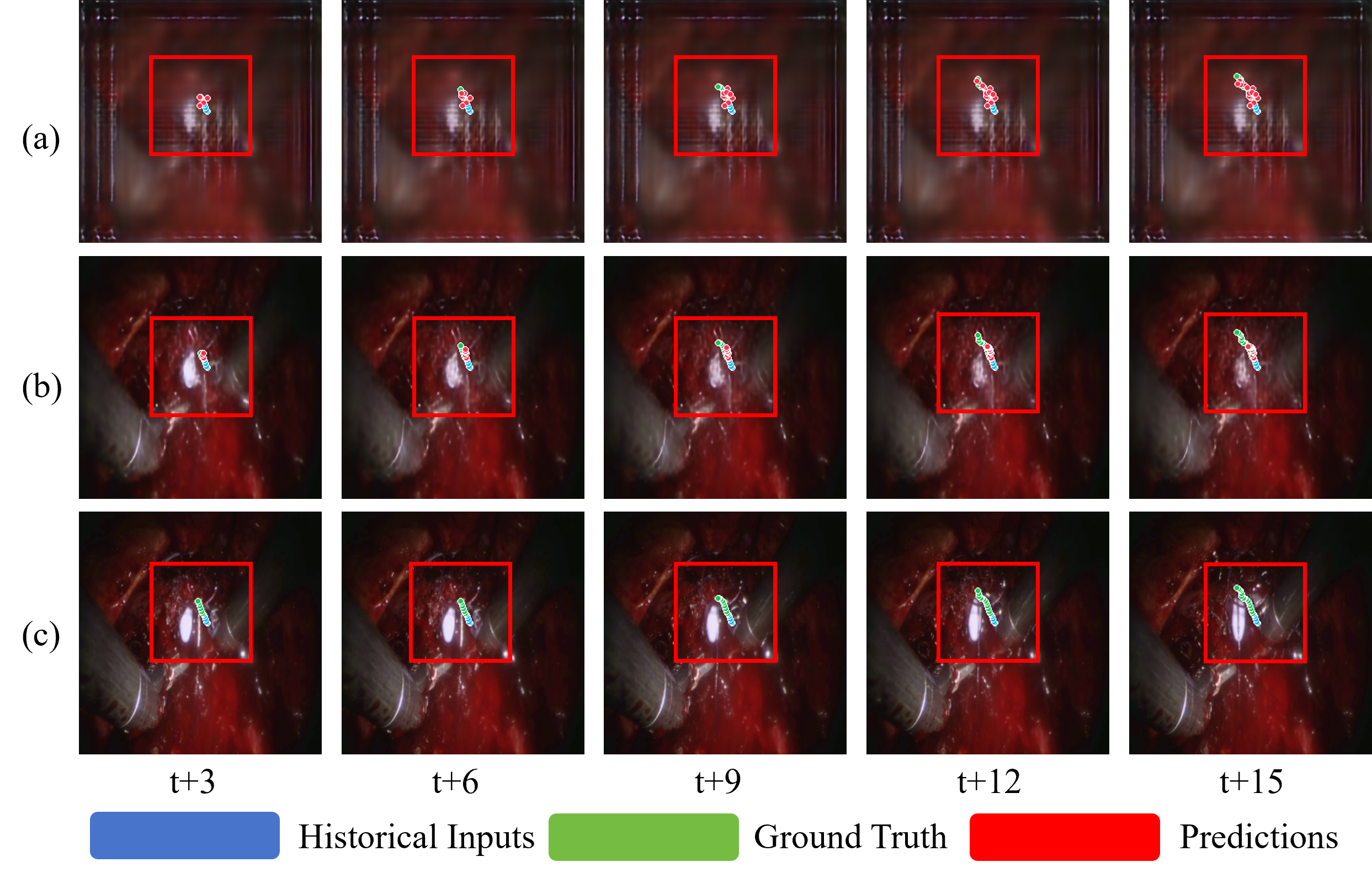}
    \caption{Qualitative comparison of direct one-shot prediction and chunked 
    autoregressive rollout. Columns show the future states at $t+3$, $t+6$, $t+9$, $t+12$, and $t+15$. Rows show the 
     (a) direct one-shot prediction, (b) chunked autoregressive rollout and (c) ground truth. Instrument trajectories are 
    overlaid on the corresponding visual states. For visualizaton purpose, we cropped the region of interest for comparision.}
    \label{fig:qualitative_results}
\end{figure}
The largest performance difference is observed in the first prediction 
segment. Compared with direct one-shot prediction, chunked rollout improves 
PSNR from 18.859 to 23.105\,dB and SSIM from 0.4713 to 0.7971. It also reduces 
LPIPS by 70.5\%, from 0.6045 to 0.1783. For trajectory prediction, the ADE and 
FDE are reduced from 45.77 and 49.47 pixels to 22.22 and 31.64 pixels, 
respectively.

The advantage of chunked prediction remains observable at longer horizons. 
For the final $t+13{:}15$ segment, chunked rollout achieves a PSNR of 
17.982\,dB and an SSIM of 0.6343, compared with 16.872\,dB and 0.4264 for 
direct prediction. Its LPIPS is also substantially lower, at 0.3047 compared 
with 0.6244. In terms of motion prediction, chunked rollout reduces ADE from 
195.27 to 158.52 pixels and FDE from 208.04 to 172.12 pixels, corresponding 
to relative reductions of approximately 18.8\% and 17.3\%, respectively.

These results suggest that decomposing a long prediction horizon into shorter 
local transitions facilitates both visual-state modeling and trajectory 
prediction. Nevertheless, both visual quality and trajectory accuracy 
gradually deteriorate as the rollout horizon increases. Therefore, 
the chunked strategy reduces, but does not eliminate, long-horizon visual 
degradation and trajectory error accumulation.

\subsection{Qualitative Results}
Figure~\ref{fig:qualitative_results} presents qualitative comparisons between 
direct one-shot prediction and chunked autoregressive rollout. We visualize 
the predictions at $t+3$, $t+6$, $t+9$, $t+12$, and $t+15$. Ground-truth future frames and 
trajectories are included as references. More qualitative results are available in the anonymized repository \url{https://github.com/anonymous-authors-code/MICCAI2026_Workshop_EndoLINA_TSWAM_Qualitative_Results.git}.

Consistent with the quantitative results, chunked rollout generally preserves 
the operative-scene structure more effectively and produces instrument 
trajectories that remain closer to the ground truth, particularly at early 
and intermediate prediction horizons. In comparison, direct one-shot 
prediction exhibits earlier visual degradation and larger trajectory 
deviations as the prediction horizon increases. At $t+12$ and $t+15$, 
however, both settings show reduced visual fidelity and noticeable endpoint 
errors. These examples further illustrate that short-chunk prediction improves 
rollout stability, while accurate long-horizon joint visual-motion forecasting 
remains challenging.

\section{Conclusion}


We present a preliminary surgical world-action model for joint future visual-state and instrument-trajectory prediction. A chunked 3→3 autoregressive rollout improves visual quality and trajectory accuracy over one-shot prediction on SurgWMBench, enabling 15-step forecasts. Performance still degrades over long horizons due to accumulated errors; future work will explore action conditioning, uncertainty modeling, and stronger visual-motion consistency.

\subsubsection*{Acknowledgments.}
This work was supported by the University of Macau under Grants SRG2024-00056-FST, 0078/2024/RIB2, and FST/SP01/2024, the Dr. Stanley Ho Medical Development Foundation under Grant SHMDF-AI/2026/ \\8 
002 and the Science and Technology Development Fund of Macao S.A.R (FDCT) 0028/2023/RIA1.

\subsubsection*{Disclosure of Interests.}
The authors have no competing interests.
%
%
%
\bibliographystyle{splncs04}
\bibliography{ref}

@article{hansen2026imitatecholec,
  title={ImitateCholec: A Multimodal Dataset for Long-Horizon Imitation Learning in Robotic Cholecystectomy},
  author={Hansen, P. and Kim, J. W. B. and Goldenberg, Anna and Chen, J. T. and Li, Y. A. and Deguet, Anton and others},
  journal={Scientific Data},
  year={2026}
}

@inproceedings{koju2025surgicalvision,
  title={Surgical Vision World Model},
  author={Koju, S. and Bastola, S. and Shrestha, P. and Amgain, S. and Shrestha, Y. R. and Poudel, R. P. K. and Bhattarai, B.},
  booktitle={MICCAI Workshop / Springer Proceedings},
  year={2025}
}

@article{rapuri2026saw,
  title={SAW: Toward a Surgical Action World Model via Controllable and Scalable Video Generation},
  author={Rapuri, S. and Seenivasan, L. and Schneider, D. and Soberanis-Mukul, R. and He, Y. and Ding, H. and others},
  journal={arXiv preprint arXiv:2603.13024},
  year={2026}
}

@article{attanasio2021autonomy,
  title={Autonomy in surgical robotics},
  author={Attanasio, Aleks and Scaglioni, Bruno and De Momi, Elena and Fiorini, Paolo and Valdastri, Pietro},
  journal={Annual Review of Control, Robotics, and Autonomous Systems},
  volume={4},
  number={1},
  pages={651--679},
  year={2021},
  publisher={Annual Reviews}
}

@inproceedings{shi2022recognition,
  title={Recognition and prediction of surgical gestures and trajectories using transformer models in robot-assisted surgery},
  author={Shi, Chang and Zheng, Yi and Fey, Ann Majewicz},
  booktitle={2022 IEEE/RSJ International Conference on Intelligent Robots and Systems (IROS)},
  pages={8017--8024},
  year={2022},
  organization={IEEE}
}

@inproceedings{biagini2025hierasurg,
  title={Hierasurg: Hierarchy-aware diffusion model for surgical video generation},
  author={Biagini, Diego and Navab, Nassir and Farshad, Azade},
  booktitle={International Conference on Medical Image Computing and Computer-Assisted Intervention},
  pages={310--319},
  year={2025},
  organization={Springer}
}

@inproceedings{chen2025surgsora,
  title={Surgsora: Object-aware diffusion model for controllable surgical video generation},
  author={Chen, Tong and Yang, Shuya and Wang, Junyi and Bai, Long and Ren, Hongliang and Zhou, Luping},
  booktitle={International Conference on Medical Image Computing and Computer-Assisted Intervention},
  pages={521--531},
  year={2025},
  organization={Springer}
}

@article{he2025surgworld,
  title={SurgWorld: Learning Surgical Robot Policies from Videos via World Modeling},
  author={He, Yufan and Guo, Pengfei and Xu, Mengya and Li, Zhaoshuo and Myronenko, Andriy and Imans, Dillan and Liu, Bingjie and Yang, Dongren and Gu, Mingxue and Ji, Yongnan and others},
  journal={arXiv preprint arXiv:2512.23162},
  year={2025}
}

@article{unterthiner2018towards,
  title={Towards accurate generative models of video: A new metric \& challenges},
  author={Unterthiner, Thomas and Van Steenkiste, Sjoerd and Kurach, Karol and Marinier, Raphael and Michalski, Marcin and Gelly, Sylvain},
  journal={arXiv preprint arXiv:1812.01717},
  year={2018}
}

@article{psychogyios2023sar,
  title={Sar-rarp50: Segmentation of surgical instrumentation and action recognition on robot-assisted radical prostatectomy challenge},
  author={Psychogyios, Dimitrios and Colleoni, Emanuele and Van Amsterdam, Beatrice and Li, Chih-Yang and Huang, Shu-Yu and Li, Yuchong and Jia, Fucang and Zou, Baosheng and Wang, Guotai and Liu, Yang and others},
  journal={arXiv preprint arXiv:2401.00496},
  year={2023}
}

@article{liu2026sutureagent,
  title={SutureAgent: Learning Surgical Trajectories via Goal-conditioned Offline RL in Pixel Space},
  author={Liu, Huanrong and Tian, Chunlin and Jia, Tongyu and Zhou, Tailai and Liu, Qin and Gao, Yu and Ban, Yutong and Gu, Yun and Rosman, Guy and Ma, Xin and others},
  journal={arXiv preprint arXiv:2603.26720},
  year={2026}
}

@article{wu2026unisurg,
  title={UniSurg: A Video-Native Foundation Model for Universal Understanding of Surgical Videos},
  author={Wu, Jinlin and Holm, Felix and Chen, Chuxi and Wang, An and Hu, Yaxin and Ye, Xiaofan and Zang, Zelin and Xu, Miao and Zhou, Lihua and Liao, Huai and others},
  journal={arXiv preprint arXiv:2602.05638},
  year={2026}
}

@article{saeidi2022autonomous,
  title={Autonomous robotic laparoscopic surgery for intestinal anastomosis},
  author={Saeidi, Hamed and Opfermann, Justin D and Kam, Michael and Wei, Shuwen and L{\'e}onard, Simon and Hsieh, Michael H and Kang, Jin U and Krieger, Axel},
  journal={Science robotics},
  volume={7},
  number={62},
  pages={eabj2908},
  year={2022},
  publisher={American Association for the Advancement of Science}
}

@article{cho2024surgen,
  title={Surgen: Text-guided diffusion model for surgical video generation},
  author={Cho, Joseph and Schmidgall, Samuel and Zakka, Cyril and Mathur, Mrudang and Kaur, Dhamanpreet and Shad, Rohan and Hiesinger, William},
  journal={arXiv preprint arXiv:2408.14028},
  year={2024}
}

@article{bengio2015scheduled,
  title={Scheduled sampling for sequence prediction with recurrent neural networks},
  author={Bengio, Samy and Vinyals, Oriol and Jaitly, Navdeep and Shazeer, Noam},
  journal={Advances in neural information processing systems},
  volume={28},
  year={2015}
}

@article{huynh2008scope,
  title={Scope of validity of PSNR in image/video quality assessment},
  author={Huynh-Thu, Quan and Ghanbari, Mohammed},
  journal={Electronics letters},
  volume={44},
  number={13},
  pages={800--801},
  year={2008},
  publisher={IET}
}

@article{wang2004image,
  title={Image quality assessment: from error visibility to structural similarity},
  author={Wang, Zhou and Bovik, Alan C and Sheikh, Hamid R and Simoncelli, Eero P},
  journal={IEEE transactions on image processing},
  volume={13},
  number={4},
  pages={600--612},
  year={2004},
  publisher={IEEE}
}

@inproceedings{zhang2018unreasonable,
  title={The unreasonable effectiveness of deep features as a perceptual metric},
  author={Zhang, Richard and Isola, Phillip and Efros, Alexei A and Shechtman, Eli and Wang, Oliver},
  booktitle={Proceedings of the IEEE conference on computer vision and pattern recognition},
  pages={586--595},
  year={2018}
}

@inproceedings{gupta2018social,
  title={Social gan: Socially acceptable trajectories with generative adversarial networks},
  author={Gupta, Agrim and Johnson, Justin and Fei-Fei, Li and Savarese, Silvio and Alahi, Alexandre},
  booktitle={Proceedings of the IEEE conference on computer vision and pattern recognition},
  pages={2255--2264},
  year={2018}
}

@article{liu2026surgwmbench,
  title={SurgWMBench: A Vision-Based Benchmark for World-Modeling Surgical Instrument Motion Planning},
  author={Liu, Huanrong and Huang, Weiliang and Zhang, Bob and Cai, Weichao and Tian, Chunlin and Li, Qingbiao},
  journal={arXiv preprint arXiv:2608.08070},
  year={2026}
}
%




\end{document}